\documentclass[conference]{IEEEtran}
\usepackage{cite}

\ifCLASSINFOpdf
  \usepackage[pdftex]{graphicx}
\else
  \usepackage[dvips]{graphicx}
\fi
\usepackage{amsmath,amssymb}
\begin{document}
%
\title{Engineering Safety in Machine Learning}

\author{\IEEEauthorblockN{Kush R. Varshney}
\IEEEauthorblockA{Mathematical Sciences and Analytics Department\\
IBM Thomas J. Watson Research Center\\
Yorktown Heights, New York 10598\\
Email: krvarshn@us.ibm.com}}


%


\maketitle

\begin{abstract}
Machine learning algorithms are increasingly influencing our decisions and interacting with us in all parts of our daily lives.  Therefore, just like for power plants, highways, and myriad other engineered sociotechnical systems, we must consider the safety of systems involving machine learning.  In this paper, we first discuss the definition of safety in terms of risk, epistemic uncertainty, and the harm incurred by unwanted outcomes.  Then we examine dimensions, such as the choice of cost function and the appropriateness of minimizing the empirical average training cost, along which certain real-world applications may not be completely amenable to the foundational principle of modern statistical machine learning: empirical risk minimization.  In particular, we note an emerging dichotomy of applications: ones in which safety is important and risk minimization is not the complete story (we name these Type A applications), and ones in which safety is not so critical and risk minimization is sufficient (we name these Type B applications).  Finally, we discuss how four different strategies for achieving safety in engineering (inherently safe design, safety reserves, safe fail, and procedural safeguards) can be mapped to the machine learning context through interpretability and causality of predictive models, objectives beyond expected prediction accuracy, human involvement for labeling difficult or rare examples, and user experience design of software.
\end{abstract}


%
\IEEEpeerreviewmaketitle

\section{Introduction}
\label{sec:intro}

In recent years, machine learning algorithms have started influencing every part of our lives, including health and wellness, law and order, commerce, entertainment, finance, human capital management, communication, transportation, and philanthropy.  As the algorithms, the data on which they are trained, and the models they produce are getting more powerful and more ingrained in society, questions about \emph{safety} must be examined.  It may be argued that machine learning systems are simply tools, that they will soon have a general intelligence that surpasses human abilities, or something in-between, but from all perspectives, they are technological components of larger sociotechnical systems that may have to be engineered with safety in mind  \cite{Conn2015}.

Safety is a commonly used term across engineering disciplines connoting the absence of failures or conditions that render a system dangerous \cite{Ferrell2010}, cf.\ safe food and water, safe vehicles and highways, safe medical treatments, safe toys, safe neighborhoods, and safe industrial plants.  Each of the domains has specific design principles and regulations that are applicable only to it; only a few works in the literature attempt a precise definition applicable to a broad set of domains and systems \cite{Moller2012}.  

In particular, a general definition of safety is the minimization of \emph{risk} and \emph{epistemic uncertainty} (understood in the usual decision-theoretic senses of the words) associated with unwanted outcomes that are severe enough to be seen as \emph{harmful} \cite{Moller2012}.  The epistemic uncertainty part of the definition is key, because harmful outcomes often occur in regimes and operating conditions that are rare, unexpected, or underdetermined.  The cost magnitude of unwanted outcomes is also key, because safety is not concerned with reducing undesired outcomes of an inconsequential nature.

With such a definition of safety, it is possible to consider domains that do not have existing safety principles and regulations such as machine learning.  The first contribution of this work is to critically examine the foundational statistical machine learning principles of empirical risk minimization and structural risk minimization \cite{Vapnik1992} from the perspective of safety.  We discuss how they, as their names imply, do not deal with epistemic uncertainty.  Furthermore, the principles rely on average losses and laws of large numbers-type arguments, which may not necessarily be fully applicable when considering safety.  Moreover, the loss functions involved in the formulations are abstract distortions between true and predicted values rather than application-specific quantities measuring loss of life, loss of quality of life, etc.\ that can be judged harmful or not \cite{Wagstaff2012}.  To the best of our knowledge, there is no existing work on analyzing machine learning using precise decision-theoretic definitions of safety.

A second contribution of this paper emerges from examining safety in formulating machine learning problems.  We find that applications of machine learning systems cluster into two types: (A) applications in which model predictions are used to support consequential decisions that can have a profound effect on people's lives, and (B) applications in which model predictions are used in settings of low consequence and large scale.  Type A applications are the ones in which safety is paramount.  We have previously noted the dichotomy of Type A and Type B applications of machine learning and data science in \cite{Varshney2015}, but did not pose them as consequences of safety definitions.  The related literature is cited in \cite{Varshney2015}, but again, does not stem from safety.

The final contribution of the paper is the discussion of strategies to increase the safety of sociotechnical systems with machine learning components.  Four categories of approaches have been identified for promoting safety in general \cite{MollerH2008}: inherently safe design, safety reserves, safe fail, and procedural safeguards.  We find and discuss examples of all of these approaches specifically for machine learning algorithms and especially to mitigate epistemic uncertainty.  Through this contribution, we can recommend strategies to engineer safer machine learning methods and set an agenda for further machine learning safety research.

The remainder of the paper is organized in the following manner.  In Section~\ref{sec:def_safety}, we discuss harm, risk, uncertainty and the definition of safety.  In Section~\ref{sec:ml}, we examine statistical machine learning from the safety perspective.  Section~\ref{sec:dich} sets forth two types of machine learning applications distinguished by their relationship to safety.  Section~\ref{sec:strat} describes ways of achieving safety in general and their specializations to machine learning.  Section~\ref{sec:conclusion} concludes.

\section{Definition of Safety}
\label{sec:def_safety}

The term \emph{safety} can have many different technical and non-technical meanings, but for our purposes, we would like to work with a precise, domain-agnostic definition.  As well-described in \cite{MollerH2008,Moller2012} and numerous references therein, such a definition of safety begins with outcomes and events.  A system yields an outcome based on its state and the inputs it receives; the outcome event may be desired or undesired.  Single events and sets of events have associated costs that can be measured and quantified by society (sometimes with difficulty).  A numeric level of morbidity, for example, can be the cost of an outcome.  An undesired outcome is only a harm if its cost exceeds some threshold.  Unwanted events of small severity are not counted as safety issues.

The next step in defining safety is to bring in decision theory and the concepts of risk and epistemic uncertainty.  Risk is the expected value of the cost of harm: we do not know what the outcome will be, but its distribution is known and we can calculate the expectation of its cost.  With uncertainty, we still do not know what the outcome will be, but in contrast to risk, its probability distribution is also unknown (or only partially known).  Epistemic uncertainty, in contrast to aleatoric uncertainty, results from lack of knowledge that could be obtained in principle, but may be practically intractable to gather.  Some decision theorists argue that all uncertainty can be captured probabilistically, but we maintain the distinction between risk and uncertainty herein, following \cite{Moller2012}.

Safety is the reduction or minimization of risk and uncertainty of harmful events.

Tomes can be and are written on costs, risk, and uncertainty.  More mathematical precision can also be given.  For our purposes, the key points in the definition of safety are that: costs have to be sufficiently high in some human sense for events to be harmful, and that safety involves reducing both the probability of expected harms and the possibility of unexpected harms.

\section{Safety and Machine Learning}
\label{sec:ml}

The starting point in the theory and practice of statistical machine learning is risk minimization.  Given joint random variables $X \in \mathcal{X}$ (features) and $Y \in \mathcal{Y}$ (labels) with probability density function $f_{X,Y}(x,y)$, a function mapping $h \in \mathcal{H}: \mathcal{X} \rightarrow \mathcal{Y}$, and a loss function $L : \mathcal{Y} \times \mathcal{Y} \rightarrow \mathbb{R}$, the risk $R(h)$ is the expectation $\mathbb{E}[L(h(X),Y)]$ = $\int_\mathcal{X}\int_\mathcal{Y}L(h(x),y)f_{X,Y}(x,y)dydx.$  The loss function $L$ typically measures the discrepancy between the value predicted for $y$ using $h(x)$ and $y$ itself, for example $(h(x)-y)^2$ in regression problems.  We would like to find the function $h$ that minimizes the risk.

However, in the machine learning context, we do not have access to the probability $f_{X,Y}$, but rather to a training set of samples drawn i.i.d.\ from the joint distribution of $X$ and $Y$: $\{(x_1,y_1),\ldots,(x_m,y_m)\}$.  The empirical risk $R^{emp}_m(h)$ is $\frac{1}{m}\sum_{i=1}^m L(h(x_i),y_i)$. The empirical risk minimization principle formulates the learning of $h$ as the minimization of $R^{emp}_m(h)$ \cite{Vapnik1992}.  Appealing to the results of Glivenko and Cantelli in empirical process theory, it can be shown that the empirical risk $R^{emp}_m(h)$ converges to the risk $R(h)$ uniformly for all $h$ as $m$ goes to infinity.  When $m$ is small (in comparison to a suitably defined complexity measure on $\mathcal{H}$), minimizing $R^{emp}_m(h)$ may not yield an $h$ that has small $R(h)$.  The structural risk minimization principle alleviates this problem by restricting the complexity of $\mathcal{H}$ or introducing regularization in the minimization problem for $h$ based on some inductive bias.

The risk minimization approach to machine learning has many strengths, as evidenced by the innumerable applied successes it has brought, and captures the risk component of safety.  However, it does not capture issues related to uncertainty and loss functions that are relevant for safety.  First, although it is assumed that the training samples $\{(x_1,y_1),\ldots,(x_m,y_m)\}$ are drawn from the true underlying probability distribution of $(X,Y)$, that may not always be the case.  Furthermore, it may be that the distribution the samples actually come from cannot be known, precluding the use of covariate shift and domain adaptation techniques.  This is one form of epistemic uncertainty that is quite relevant to safety because training on a data set from a different distribution can cause much harm.

Also, it may be that the training samples do come from the true, but unknown, underlying distribution, but are absent from large parts of the $\mathcal{X}\times\mathcal{Y}$ space due to small probability density there.  Here the learned $h$ will be completely dependent on the inductive bias rather than the uncertain true distribution, which could introduce a safety hazard.

As mentioned above, statistical learning theory analysis utilizes laws of large numbers to study the effect of finite training data and the convergence of the empirical risk to the true risk, but in considering safety, we should also be cognizant that in deployment, a machine learning system only encounters a finite number of test samples and the actual operational risk is an empirical quantity on the test set.  Thus the operational risk may be much larger than the true risk for small cardinality test sets, even if $h$ is risk-optimal.  This uncertainty caused by the instantiation of the test set can have large safety implications on individual test samples.

As we discussed above, the domain of the loss function in risk minimization is $\mathcal{Y}\times\mathcal{Y}$ and the output is an abstract quantity representing prediction error.  However, in real-world applications, the value of the loss function may be endowed with some human cost and that human cost may imply a loss function that also includes $\mathcal{X}$ in the domain.  Moreover, the cost may be severe enough to be harmful and thus a safety issue in some parts of the domain and not in others.  

\section{Type A and Type B Applications}
\label{sec:dich}

Having described general considerations for machine learning in terms of safety in Section~\ref{sec:ml}, we examine safety considerations in specific applications of machine learning systems in this section.  

\subsection{Harmful Costs}

We begin with the severity of unwanted outcomes.  Predictions made by machine learning systems in applications such as medical diagnosis \cite{FosterKS2014}, loan approval \cite{LessmannSBT2015}, and prison sentencing \cite{BerkH2015} can have a profound effect on people; undesired outcomes are truly harmful in a human sense.  In contrast, other applications of machine learning are of a less consequential nature; examples include streaming services deciding on the compression level of video packets to transmit to subscribers every few seconds \cite{EisingerRG2008,Govind2014}, web portals deciding which news story to show on top \cite{AgarwalCER2013}, and speech transcription systems classifying phonemes \cite{HintonDYDMJSVNSK2012}.  The quality of service implications of unwanted outcomes in such applications are not safety hazards.

Taking a more nuanced look at costs and undesirable predictions, we note that loss functions are not always monotonic in the correctness of predictions and depend on whose perspective is in the objective.  Consider the loan approval application: the applicant would like an approval decision regardless of their features indicating ability to repay, the lender would like approval only in cases in which applicant features indicate likely repayment, and society would like there to be fairness or equitability in the system so that protected groups, such as defined by race and gender, are not discriminated against.  The lender perspective is consistent with the typical choice of loss function, but the others are not.

\subsection{Epistemic Uncertainty}

The type of machine learning applications with potentially harmful consequences and the type with harmless consequences can be further analyzed with respect to epistemic uncertainty.  There is no a priori reason for the applications to follow the same type structure when examining uncertainty, but as we discuss in the following, the two types are recapitulated.  For ease of reference, let the medical diagnosis, loan approval, prison sentencing-type applications constitute Type A, and the other class of applications Type B.  This is the same nomenclature as in \cite{Varshney2015}.

In addition to the lack of severity of costs, another characteristic of Type B applications is that they are performed at scales with large training sets, large testing sets, and the ability to explore the feature space.  For example, in the web portal news story application, one can use billions of data points as training, perform large-scale A/B testing, and evaluate average performance on millions or billions of clicks.  For these reasons, the epistemic uncertainties discussed in Section~\ref{sec:ml} are less prevalent in Type B applications than in Type A applications.  In contrast, in Type A applications it is more often than not the case that there is uncertainty about the training samples being representative of the testing samples, and that only a few predictions are made.  Uncertainty of the various types discussed is common in Type A applications.

Thus, not only are errors in Type B applications less costly in human terms, but the amount of uncertainty in the system is less.  Therefore, for both reasons, safety is much less relevant in Type B applications than Type A applications.  The focus in Type B applications can be squarely on risk minimization whereas Type A applications require the consideration of strategies for achieving safety, as we discuss next.

\section{Strategies for Achieving Safety}
\label{sec:strat}

As discussed in the introduction, safety is usually investigated on an application-by-application basis and strategies for achieving it the same.  For example, setting the minimum thickness of vessels and removing flammable materials from a chemical plant are ways of achieving safety.  Analyzing such strategies across domains, \cite{MollerH2008} has identified four main categories of approaches to achieve safety.  In this section, we discuss each of these categories in turn along with specific approaches that extend machine learning formulations beyond risk minimization for safety.

\subsection{Inherently Safe Design}
\label{sec:strat:inherent}

Inherently safe design is the exclusion of a potential hazard from the system (instead of controlling the hazard).  For example, excluding hydrogen from the buoyant material of a dirigible airship makes it safe.  (Another possible safety measure would be to introduce apparatus to prevent the hydrogen from igniting.)

In the machine learning context, we would like robustness against the uncertainty of the training set not being sampled from the test distribution.  The training set may have various quirks or biases that are unknown to the user and that will not be present during the test phase.  Highly complex modeling techniques used today, including extreme gradient boosting and deep neural networks, may pick up on those data vagaries in the learned models they produce to achieve high accuracy, but might fail due to an unknown shift in the data domain.  

The models are so complex that it is very difficult to understand how they will react to such shifts and whether they will produce harmful outcomes as a result.  Two related ways to introduce inherently safe design is by insisting on models that can be interpreted by people and by excluding features that are not causally-related to the outcome \cite{Welling2015}.  By examining interpretable models, features or functions capturing quirks in the data can be noted and excluded, thereby avoiding related harm.  Similarly, by excluding non-causal variables, phenomena that are not a part of the true `physics' of the system can be excluded and related harm avoided.  

The desire for neither interpretability nor causality of models is captured in the standard risk minimization formulation of machine learning.  Extra regularization or constraints on $\mathcal{H}$ beyond those implied by structural risk minimization are needed to learn such models.  There may be performance loss in accuracy by doing so when measuring accuracy with a common training and testing data probability distribution, but the reduction in epistemic uncertainty by doing so increases safety.  Both interpretability and causality may be incorporated into a single learned model, e.g.~\cite{WangR2015b}.

\subsection{Safety Reserves}
\label{sec:strat:reserves}

A second strategy for achieving safety is through multiplicative or additive reserves, known as safety factors and safety margins, respectively.  In mechanical systems, a safety factor is a ratio between the maximal load that does not lead to failure and the load for which the system was designed.  Similarly the safety margin is the difference between the two.

For the purposes of machine learning with uncertainty, whether that uncertainty is in the training data matching the test distribution or in the instantiation of the test set, we can parameterize the unknown with the symbol $\theta$.  Let the risk of the risk-optimal model if the $\theta$ were known be $R^*(\theta)$.  Along the same lines as safety factors and safety margins, robust formulations find $h$ while constraining or minimizing $\max_\theta \frac{R(h,\theta)}{R^*(\theta)}$ or $\max_\theta \left(R(h,\theta) - R^*(\theta)\right)$.  Such formulations can capture uncertainty in the class priors and uncertainty resulting from label noise in classification problems.  They can also capture the uncertainty of which part of the $\mathcal{X}$ space the actual small set of test samples comes from: we do not care as much about average test error for medical diagnosis problems if a model will only be used on a handful of patients as we do about the maximum test error.

A different sort of safety factor comes about when considering fairness and equitability.  In certain prediction problems, the risk of harm for members of protected groups should not be much worse (up to a multiplicative factor) than the risk of harm for others \cite{FeldmanFMSV2015}.  Features indicating a protected group, such as race and gender, are dimensions in the $\mathcal{X}$ space; we can partition the space into the sets $\mathcal{X}_p, \mathcal{X}_u \subset \mathcal{X}$ corresponding to the protected and unprotected groups respectively.  The safety factor known as disparate impact constrains the following to a minimum value such as $4/5$: $$\frac{\int_{\mathcal{X}_p}\int_\mathcal{Y}L(x,h(x),y)f_{X,Y}(x,y)dydx}{\int_{\mathcal{X}_u}\int_\mathcal{Y}L(x,h(x),y)f_{X,Y}(x,y)dydx}.$$  Under such a constraint, the risk of harm for protected groups is not much more than for unprotected groups.

\subsection{Safe Fail}
\label{sec:strat:fail}

The third general category of safety measures is `safe fail,' which implies that a system remains safe when it fails in its intended operation.  Examples are electrical fuses, so-called dead man's switches on trains, and safety valves on boilers.  

A technique used in machine learning when predictions cannot be given confidently is the reject option \cite{VarshneyPMCH2013}: the model reports that it cannot reliably give a prediction and does not attempt to do so, thereby failing safely.  When the model elects the reject option, typically a human operator intervenes, examines the test sample, and provides a manual prediction.

In classification problems, models are reported to be least confident near the decision boundary.  However, by doing so, there is an implicit assumption that distance from the decision boundary is inversely related to confidence.  This is reasonable in parts of $\mathcal{X}$ with high probability density and large numbers of training samples because the decision boundary is located where there is a large overlap in likelihood functions.  However, as discussed in Section~\ref{sec:ml}, parts of $\mathcal{X}$ with low density may not contain any training samples at all and the decision boundary may be completely based on an inductive bias, thereby containing much epistemic uncertainty.  In these parts of the space, distance from the decision boundary is fairly meaningless and the typical trigger for the reject option should be avoided \cite{AttenbergIP2015}.  For a rare combination of features in a test sample \cite{Weiss2004}, a safe fail mechanism is to always go for manual examination.

\subsection{Procedural Safeguards}
\label{sec:strat:procedure}

Finally, the fourth strategy for achieving safety is  given the name procedural safeguards.  This strategy includes measures beyond ones designed into the core functionality of the system, such as audits, training, posted warnings, and so on.  Two directions in machine learning that can be used for increasing safety within this category are user experience design and openness.  

In Type A applications especially, non-specialists are often the operators of machine learning systems.  Defining the training data set and setting up evaluation procedures, among other things, have certain subtleties that can cause harm during operation if done incorrectly.  User experience design can be used to guide and warn novice and experienced practitioners to set up machine learning systems properly and thereby increase safety.

Best of breed machine learning algorithms these days are open source, which allows for the possibility of public audit.  Safety hazards and potential harms can be discovered through examination of source code.  However, open source software is not enough, because the behavior of machine learning systems is driven by data as much as it is by software implementations of algorithms.  Open data refers to data that can be freely used, reused and redistributed by anyone.  It is more common in Type A applications such as those sponsored or run by governments than in Type B applications where the data is oftentimes the key value proposition.  Opening data is a procedural safeguard for increasing safety that is increasingly being adopted in Type A applications.

\section{Conclusion}
\label{sec:conclusion}

Machine learning systems are already embedded in many functions of society.  The prognosis is for broad adoption to only increase across all areas of life.  With this prevailing trend, machine learning researchers, engineers, and ethicists have started discussing the topic of safety.  In this paper, we contribute to this discussion starting from a very basic definition of safety in terms of harm, risk, and uncertainty and building upon it in the machine learning context.  We identify that the minimization of epistemic uncertainty is missing from standard modes of machine learning developed around risk minimization and that it needs to be included when considering safety.  We have delineated two types of applications of machine learning: Type A in which safety is an important concern and Type B in which it is not so.  We have discussed several strategies for increasing safety that are especially pertinent in Type A applications.

Within safety engineering, there is a further subdivision into the concepts of substantive safety and nominal safety.  All of the design elements in a nominally safe system meet regulations and design criteria.  Substantive safety is the long-term performance that the system actually exhibits.  We have not made this distinction in this paper, but it is worth doing so in future work.  

Also, all of the safety discussion in this paper has been related to predictions or outcomes based on predictions.  However, there are other parts of a machine learning system besides the core prediction component in which we should also consider safety.  For example, privacy and disclosure risk in microdata release is a safety issue that is not part of the main prediction model, but is part of the larger machine learning sociotechnical system \cite{Varshney2015}.  Further work should study all other parts of machine learning in a similar fashion as this work, starting from the first principles of safety in terms of cost, risk, and uncertainty.

The strategies for increasing safety that we mentioned in Section~\ref{sec:strat} are not a comprehensive list and are far from fully developed.  This paper can be seen as laying the foundations for a research agenda motivated by Type A applications and safety within which further strategies can be developed and existing strategies can be fleshed out.  In some respects, the research community has taken risk minimization close to the limits of what is achievable.  Safety, especially epistemic uncertainty minimization, represents a direction that offers new and exciting problems to pursue.  As it is said in the Sanskrit literature, \emph{ahi\d{m}s\={a} paramo dharma\d{h}} (non-harm is the ultimate direction).




\section*{Acknowledgment}

The author thanks Cynthia Rudin for asking what he means by `safety,' and Dennis Wei and Been Kim for asking what he means by `robustness.'

\newpage


\bibliographystyle{IEEEtran}
\bibliography{IEEEabrv,safety}
%



\end{document}